\documentclass[sigconf]{acmart}

\usepackage{booktabs}
\usepackage{amsmath}
\usepackage{array}
\usepackage{tikz}

\AtBeginDocument{%
  }

\copyrightyear{2026}
\acmYear{2026}
\setcopyright{cc}
\setcctype{by}
\acmConference[MM '26]{Proceedings of the 34th ACM International Conference on Multimedia}{November 10--14, 2026}{Rio de Janeiro, Brazil}
\acmBooktitle{Proceedings of the 34th ACM International Conference on Multimedia (MM '26), November 10--14, 2026, Rio de Janeiro, Brazil}
\acmDOI{10.1145/3767308.3837697}

\acmISBN{979-8-4007-2213-4/2026/11}

\begin{document}

\title{EEGBind: Detecting Source-Level Interictal Epileptiform Discharges via EEG-Centric Multimodal Binding}

\author{Muchen Li}
\orcid{0009-0000-0142-0865}
\affiliation{%
  \department{Artificial Intelligence Thrust}
  \institution{The Hong Kong University of Science and Technology (Guangzhou)}
  \city{Guangzhou}
  \state{Guangdong}
  \country{China}}
\email{mli893@connect.hkust-gz.edu.cn}

\author{Anglin Liu}
\orcid{0009-0003-3519-2655}
\affiliation{%
  \department{Artificial Intelligence Thrust}
  \institution{The Hong Kong University of Science and Technology (Guangzhou)}
  \city{Guangzhou}
  \state{Guangdong}
  \country{China}}
\email{aliu104@connect.hkust-gz.edu.cn}

\author{Xuetian Gao}
\orcid{0009-0000-8577-5187}
\affiliation{%
  \institution{Ringgee Smart Technologies Co., Ltd.}
  \city{Fuzhou}
  \state{Fujian}
  \country{China}}
\email{gaoxuetian@ringgee.com}

\author{Ruijian Xu}
\orcid{0000-0002-3757-0387}
\affiliation{%
  \institution{Ringgee Smart Technologies Co., Ltd.}
  \city{Fuzhou}
  \state{Fujian}
  \country{China}}
\email{xuruijian@ringgee.com}

\author{Jintai Chen}
\authornote{Corresponding author.}
\orcid{0000-0002-3199-2597}

\affiliation{%
  \department{Artificial Intelligence Thrust}
  \institution{The Hong Kong University of Science and Technology (Guangzhou)}
  \city{Guangzhou}
  \state{Guangdong}
  \country{China}}
\email{jintaichen@hkust-gz.edu.cn}

\begin{abstract}
Source-level analysis of interictal epileptiform discharges (IEDs) is relevant
to presurgical evaluation and treatment planning because it helps characterize
where epileptiform activity is likely to arise. Beyond detecting whether an IED
is present, this setting requires assigning IED-positive activity to clinically
meaningful brain-region categories. This setting is challenging because
source-region evidence in short electroencephalography (EEG) windows can be
subtle, partial, and affected by subject variability, class imbalance, and
imperfect multimodal context. We present EEGBind, an EEG-centric multimodal
binding framework for five-class source-level IED classification. EEGBind treats
EEG as the primary modality and
binds synchronized video-context features around an EEG-centric representation.
Instead of relying on early or overly strong multimodal fusion, which may
perturb the source-sensitive EEG representation, EEGBind uses video context as
auxiliary evidence for robust classification. A view-consistent repair stage is
further used to improve hidden-set robustness while preserving the learned
source-class boundary. On the
NeuroMM 2026 Grand Challenge Track 3 NMM-Source-IED benchmark, EEGBind achieves
0.8395 on weighted-F1 and outperforms strong competitors. These results support
EEG-centric multimodal binding as a practical strategy for source-level IED
classification. The open-source code is available at \url{https://github.com/HKUSTGZ-ML4Health-Lab/NeuroMM2026_IED_Detection}.
\end{abstract}


\maketitle

\section{Introduction}

Interictal epileptiform discharges (IEDs) provide clinically useful evidence for
supporting epilepsy diagnosis, classification, and localization
\cite{noachtar2009role}. In clinical review, detecting whether an epileptiform
event is present is only one part of the analysis. It is also important to
characterize where epileptiform activity is likely to arise, because
source-level evidence can support presurgical evaluation, treatment planning, and
the interpretation of focal or generalized epileptiform patterns. Automated
source-level IED analysis is therefore a meaningful extension of binary IED
detection: the system must assign an IED-positive candidate to a clinically
meaningful source-region category rather than only decide whether the candidate
is epileptiform.

This source-level setting is technically challenging. Regional evidence in short
electroencephalography (EEG) windows can be subtle, partial, and distributed
across channels. Similar transient morphologies may appear with different
spatial expression across subjects, recording conditions, and source regions
\cite{lin2025vepiset}. As a result, direct EEG-only source classification may
produce a useful representation but still struggle to separate generalized,
frontal, temporal, occipital, and centro-parietal categories. The difficulty is
different from score-based IED detection: the model must preserve
source-discriminative structure among already IED-positive candidates, not only
rank candidate windows by the likelihood of containing an IED.

We study this problem on the NMM-Source-IED benchmark. Defined in Track 3 of the
NeuroMM 2026 Grand Challenge~\cite{neuromm2026challenge}, this patient-disjoint
multimodal benchmark provides IED-positive candidate windows with synchronized
EEG and video context and requires each candidate to be assigned to one of five
source-region classes. Performance is measured by weighted-F1 on patients unseen
during training. This standardized setting isolates a source-level decision that
is often hidden inside broader EEG interpretation pipelines and enables
controlled study of whether multimodal context improves source-level
classification without turning the problem into unconstrained video--EEG fusion.

Synchronized video context can reduce candidate-level ambiguity, but it also
creates a modality-role mismatch. Visual context can describe patient motion,
posture, recording state, or other candidate-level conditions that are not fully
captured by EEG alone
\cite{peh2022artifact,zhang2024wearablemultimodal,lin2024vepinet}. However, the
source-region label remains primarily tied to electrophysiological morphology.
Therefore, video should provide bounded candidate context rather than act as an
independent source classifier. This differs from generic multimodal fusion, and
is closer to a modality-centric binding problem: one modality organizes the
representation while auxiliary modalities contribute contextual evidence.
Language-centered multimodal binding follows a related principle by using
language to organize multiple auxiliary modalities~\cite{zhu2024languagebind};
for NMM-Source-IED, EEG is the appropriate organizing modality because the
target label is source-level electrophysiological activity.

Motivated by these observations, we present EEGBind, an EEG-centric multimodal
binding framework for source-level IED classification. At its core, the
binding mechanism uses ST-EEGFormer as the EEG backbone and binds
synchronized video-context routes around an EEG-derived source representation.
A CE-free view-consistent repair stage further improves robustness under
unstable candidate context. On the benchmark's official hidden-test evaluation,
EEGBind achieves a weighted-F1 of 0.8395 and ranks first among all participating
methods. 

Our contributions are summarized as follows:
\begingroup
\makeatletter
\let\EEGBindDefaultListI\@listi
\def\@listi{\EEGBindDefaultListI
  \leftmargin=1.4em
  \topsep=4pt
  \itemsep=1pt
  \parsep=0pt\relax}
\makeatother
\begin{itemize}
  \item We propose EEGBind for EEG-centric multimodal binding in source-level IED
  classification. Its binding mechanism uses EEG as the source
  anchor and synchronized video routes as bounded context, addressing the
  modality-role mismatch between electrophysiological and visual evidence.

  \item We introduce a CE-free view-consistent candidate repair strategy that
  refines multimodal representations under route-dropped and masked video
  contexts without further moving the learned source-class boundary. This
  reliability-oriented repair improves robustness to unstable candidate
  context while preserving source-discriminative structure.

  \item We conduct systematic benchmark evaluation and controlled studies
  showing that the gain of EEGBind comes from meaningful video-context binding
  rather than additional model capacity. The final subject-disjoint five-fold
  ensemble achieves a weighted-F1 of 0.8395 and ranks first on the official
  NMM-Source-IED evaluation.
\end{itemize}
\endgroup

\section{Related Work}

\subsection{IED Analysis and Source-Level Classification}

Automatic IED analysis has traditionally focused on detecting epileptiform
events in EEG recordings, a modality central to clinical epilepsy
evaluation~\cite{noachtar2009role}. Earlier approaches used
morphology-sensitive signal processing, template matching, wavelet features, and
other hand-designed descriptors to capture sharp epileptiform transients
\cite{latka2003wavelet,rosado2016automatic}. Modern EEG foundation models
provide stronger transferable representations and adaptation across acquisition
setups for downstream EEG tasks
\cite{jiang2024labram,chen2024eegformer,yang2026steegformer,elouahidi2026reve}.
Recent work has
also explored cross-representation modeling for EEG. EEG-VL converts EEG
signals into visual representations, extracts features with a pretrained image
encoder, and conditions a language model for seizure
detection~\cite{liang2025eegvl}. Unlike EEG--video methods, its visual features
are derived from EEG itself rather than from synchronized patient video. The
NeuroMM source-classification setting extends this detection-oriented
view~\cite{lin2025vepiset,neuromm2026challenge}: the system must distinguish
source-related IED categories rather than simply separate IED from non-IED
windows, making the problem more sensitive to subject variation, class
imbalance, and local morphology ambiguity.

\subsection{Multimodal Fusion and EEG-Centric Binding}

Clinical EEG interpretation can benefit from contextual evidence beyond the EEG
waveform. Auxiliary physiological signals and visual observations may reveal
patient movement, posture, muscle activation, and recording
artifacts~\cite{peh2022artifact,zhang2024wearablemultimodal}. vEpiNet combines
EEG representations with video-derived motion and pose cues for binary IED
detection and has been evaluated in a prospective clinical
setting~\cite{lin2024vepinet}. EEGVFusion further integrates self-supervised EEG
pretraining, spatiotemporal video encoding, optimal-transport alignment, and
bidirectional cross-attention for seizure detection~\cite{lu2026eegvfusion}.
Although these EEG--video approaches have shown promising results in binary
event detection, their application to the more fine-grained problem of
source-level IED classification remains underexplored. Video primarily captures
externally observable information, such as patient posture and movement, which
can help identify motion-related artifacts and distinguish epileptic events from
non-IED activity. However, such observations are only indirectly related to the
neurophysiological morphology that determines the IED source, while unrelated
movements may introduce misleading context. Addressing this gap therefore
requires a multimodal design that can exploit useful video context without
overriding source-sensitive EEG representations. EEGBind follows this principle
by placing EEG at the center of source-level classification and incorporating
video features as auxiliary contextual evidence.

\subsection{Robust Multimodal Learning}

Robust multimodal learning addresses incomplete inputs in several ways. MARIA
applies mask-aware attention only to available clinical inputs, without explicit
imputation~\cite{caruso2025maria}; TRML infers virtual representations for
missing modalities and aligns their semantic spaces~\cite{zhao2024trml};
PhysioOmni separates invariant and modality-specific representations and uses
masked pretraining and resilient fine-tuning for arbitrary missing
modalities~\cite{fu2025physioomni}.

Representation consistency offers a complementary approach. Joint-embedding
objectives align perturbed views without reconstructing input
details~\cite{assran2023ijepa}, while momentum encoders and feature queues
provide stable contrastive targets~\cite{he2020moco}. These methods do not
directly address the asymmetric setting in which a consistently available
primary modality is accompanied by incomplete or unstable auxiliary routes.

\begin{figure*}[t]
  \centering
  \includegraphics[width=\textwidth, keepaspectratio]{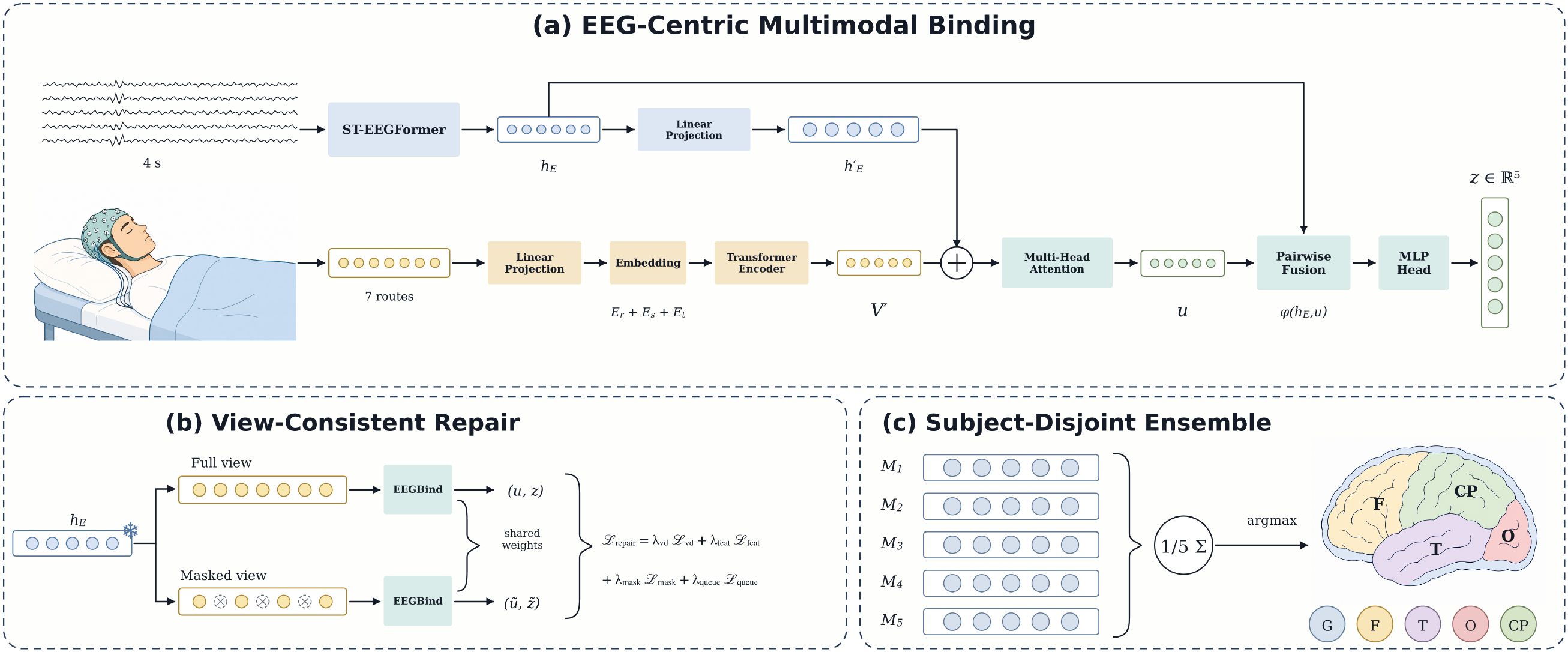}
  \caption{Overview of EEGBind for source-level IED classification:
  EEG-centric multimodal binding,
  cross-entropy-free (CE-free) view-consistent repair, and subject-disjoint
  five-fold ensembling.}
  \Description{A three-panel diagram of EEGBind. Panel (a) shows a four-second
  EEG window encoded by ST-EEGFormer and seven synchronized video-feature routes
  entering an attentive probe. An expanded view of the probe shows separate
  linear projections for EEG and video, route-aware video embeddings, a
  Transformer encoder, multi-head attention, pairwise EEG--context fusion, and
  an MLP that produces five source-class logits. Panel (b) shows CE-free
  view-consistent repair: full and masked video-route views pass through shared
  EEGBind weights, and their representations and logits are aligned using
  view-drop, feature-target, masked-route, and queue-based objectives while the
  EEG encoder remains fixed. Panel (c) shows five subject-disjoint repaired
  models whose class log probabilities are averaged before an argmax produces
  one of five source labels: generalized, frontal, temporal, occipital, or
  centro-parietal.}
  \label{fig:method_overview}
\end{figure*}

\section{Methodology}
\label{sec:methodology}

\subsection{Problem Formulation}

Figure~\ref{fig:method_overview} summarizes the overall EEGBind pipeline. We
formulate Track 3~\cite{neuromm2026challenge} as five-class source-region
classification over official IED-positive candidate windows. EEGBind follows
an EEG-centric multimodal binding design: the EEG
window provides the primary source representation, while synchronized
video-context features are used as auxiliary evidence around that
representation.

Let $i$ index an official candidate window. For Track 3, the system predicts one
of five source-region labels corresponding to the NMM-Source-IED categories.
Each candidate contains an EEG window $x_i^{\mathrm{E}}$ and synchronized
video-context features $\{x_{i,b}^{\mathrm{V}}\}_{b=1}^{B}$. The model produces
logits $z_i\in\mathbb{R}^{5}$ and predicts IED source level $\hat{y}_i \in \{1,2,3,4,5\}$, where $\hat{y}_i$ denotes the index $k$ that yields the maximum value for $z_{i,k}$. 

\subsection{EEG-Centric Multimodal Binding}

The EEG stream provides the primary representation for source classification.
Given an EEG window $x_i^{\mathrm{E}}$, the encoder produces an EEG feature,
while video-context features from multiple synchronized routes are projected
into a shared hidden space:
\begin{equation}
  h_i^{\mathrm{E}} = f_{\theta}(x_i^{\mathrm{E}}), \qquad
  h_{i,b}^{\mathrm{V}} = \phi_b(x_{i,b}^{\mathrm{V}}),\ b=1,\dots,B .
\end{equation}

An attentive probe combines the EEG representation and video-context tokens to
produce a candidate representation $u_i$ and source logits $z_i$:
\begin{equation}
  u_i = A_{\psi}\left(h_i^{\mathrm{E}},
  \{h_{i,b}^{\mathrm{V}}\}_{b=1}^{B}\right), \qquad
  z_i = g_{\omega}(u_i).
\end{equation}
The binding module is EEG-centric: video-context tokens provide auxiliary
evidence around the EEG representation, but the source prediction is produced
from the bound candidate representation rather than from an independent visual
classifier.

\subsection{CE-Free View-Consistent Repair}

After multimodal binding, we apply a short representation repair stage. The
first-stage model has already learned a supervised source-classification
boundary, so the repair stage disables supervised cross-entropy and other
boundary-moving objectives. It instead encourages stable latent and logit
behavior under missing or perturbed video-context routes.

During repair, the cached EEG representation, EEG-to-binding projection, and
five-class source classifier remain fixed, while the video-context projections
and attentive fusion modules are updated. A full-context forward pass with route
dropout disabled provides detached logit and feature targets, and an EMA copy of
the model provides keys for the queue-based objective. The perturbed branch
applies independent Bernoulli dropout to the seven video routes with probability
0.15 while retaining at least three routes. For masked-route prediction, one or
two routes are masked while at least five remain visible. All available routes
are used at inference. This configuration introduces no additional
label-supervised update to the classifier during repair.

Let $u_i$ and $z_i$ denote the full-view latent representation and logits, and
let $\tilde{u}_i$ and $\tilde{z}_i$ denote the corresponding outputs under a
route-dropped or route-masked view. The repair loss is
\begin{equation}
  \mathcal{L}_{\mathrm{repair}} =
  \lambda_{\mathrm{vd}}\mathcal{L}_{\mathrm{vd}}+
  \lambda_{\mathrm{feat}}\mathcal{L}_{\mathrm{feat}}+
  \lambda_{\mathrm{mask}}\mathcal{L}_{\mathrm{mask}}+
  \lambda_{\mathrm{queue}}\mathcal{L}_{\mathrm{queue}} .
\end{equation}
The view-drop consistency term aligns full-view and dropped-view predictions
using Kullback--Leibler (KL) divergence:
\begin{equation}
  \mathcal{L}_{\mathrm{vd}} =
  D_{\mathrm{KL}}\left(\operatorname{softmax}(z_i/T)
  \,\|\, \operatorname{softmax}(\tilde{z}_i/T)\right),
\end{equation}
where $T$ is a temperature used for soft prediction consistency.

The feature-target consistency term~\cite{assran2023ijepa} aligns the
dropped-view latent with a stopped-gradient full-view target:
\begin{equation}
  \begin{aligned}
    \mathcal{L}_{\mathrm{feat}}
    &= 1-\cos\left(\tilde{u}_i, \operatorname{sg}(u_i)\right), \\
    \mathcal{L}_{\mathrm{mask}}
    &= 1-\cos\left(u_i^{\mathrm{mask}}, \operatorname{sg}(u_i)\right).
  \end{aligned}
\end{equation}
The masked-route term applies the same cosine objective when video-context
routes are masked.

The queue-based contrastive term~\cite{he2020moco} uses a route-dropped student
query $q_i$, an exponential moving average (EMA) full-view teacher key
$k_i^{+}$, and a queue of previous keys $\mathcal{Q}$:
\begin{equation}
  \begin{aligned}
    \ell_i^{+} &= \exp(q_i^{\top}k_i^{+}/\tau), \qquad
    \ell_i(k)=\exp(q_i^{\top}k/\tau), \\
    \mathcal{L}_{\mathrm{queue}} &=
    -\log
    \frac{\ell_i^{+}}
    {\ell_i^{+}+\sum_{k\in\mathcal{Q}}\ell_i(k)} .
  \end{aligned}
\end{equation}
The queue-based contrastive objective is used as a weak latent-structure term
rather than as the main driver of source-classification learning.

\subsection{Subject-Disjoint Fold Ensemble}

For the official evaluation, EEGBind uses subject-disjoint five-fold training.
For each fold $m$, we train the same model and repair it with the CE-free
objective, producing logits $z_i^{(m)}$ for each candidate. We convert these
logits to log probabilities and average them with equal weights:
\begin{equation}
  \begin{aligned}
    \bar{\ell}_{i,k} &=
    \frac{1}{M}\sum_{m=1}^{M}
    \log \operatorname{softmax}(z_i^{(m)})_k,\\
    \hat{y}_i &= \arg\max_k \bar{\ell}_{i,k}.
  \end{aligned}
\end{equation}
where $M=5$. Using subject-disjoint folds reduces sensitivity to a particular subject split
while keeping all ensemble members tied to the same EEG-centric binding design.

\section{Experiments}

\subsection{Dataset and Evaluation Metric}

We evaluate on NMM-Source-IED, Track 3 of the NeuroMM 2026 Grand
Challenge~\cite{neuromm2026challenge}. The benchmark provides official
IED-positive candidate windows with multimodal neuro-signal context. The system
predicts one of five source-region classes for every candidate: generalized,
frontal, temporal, occipital, and centro-parietal.

The primary metric is \textit{weighted-F1}:
\begin{equation}
  \mathrm{weighted\text{-}F1} =
  \sum_{k=1}^{5}\frac{n_k}{\sum_j n_j}F1_k ,
\end{equation}
where $F1_k$ is the F1 score of class $k$ and $n_k$ is the number of evaluated
examples in that class.

\subsection{Official Evaluation Analysis}

\begin{figure}[t]
  \centering
  \includegraphics[width=\linewidth]{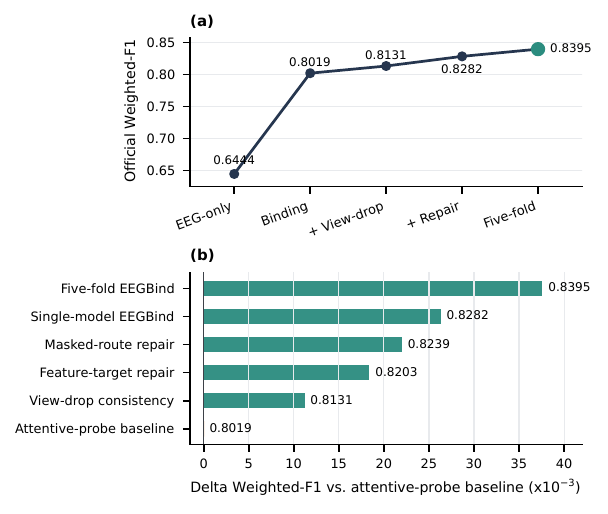}
  \caption{Official Track 3 weighted-F1 progression for the EEG-only baseline,
  EEGBind variants, and the final ensemble; Table~\ref{tab:official_results}
  reports the exact values.}
  \Description{A two-panel plot showing the official weighted-F1 trajectory and
  selected repair and aggregation variants measured against the attentive-probe
  baseline.}
  \label{fig:official_scores}
\end{figure}

Table~\ref{tab:official_results} summarizes the EEG-only baseline and evaluated
EEGBind variants, ending with the five-fold ensemble. The EEG-only
ST-EEGFormer baseline obtains 0.6444 on weighted-F1, indicating a substantial
hidden-set gap for direct EEG-only source classification. The single-model
repair sequence improves from the attentive-probe baseline to the single-model
repaired EEGBind, reaching 0.8282 on weighted-F1. The five-fold EEGBind ensemble
applies the same repair recipe in a subject-disjoint setup and reaches 0.8395 on weighted-F1. Figure~\ref{fig:official_scores} visualizes the same official score
progression, from the EEG-only baseline to the final EEGBind ensemble.

\begin{table}[t]
  \centering
  \caption{Official Track 3 Weighted-F1 results for EEGBind variants and
  post-evaluation video-context controls.}
  \label{tab:official_results}
  \begin{tabular}{@{}>{\raggedright\arraybackslash}p{0.70\linewidth}c@{}}
    \toprule
    System or control & Weighted-F1 \\
    \midrule
    EEG-only ST-EEGFormer baseline & 0.6444 \\
    Attentive-probe baseline & 0.8019 \\
    View-drop attentive-probe baseline & 0.8131 \\
    Feature-target repair variant & 0.8203 \\
    Masked-route repair variant & 0.8239 \\
    Single-model repaired EEGBind & 0.8282 \\
    Five-fold EEGBind ensemble & 0.8395 \\
    \midrule
    \multicolumn{2}{@{}l}{\textit{Post-evaluation video-context controls}} \\
    Matched-capacity dummy-video attentive probe & 0.7448 \\
    Five-fold EEGBind, zero-video inference & 0.7460 \\
    \bottomrule
  \end{tabular}
\end{table}

We additionally performed two post-evaluation controls to separate useful
video-context binding from model-capacity effects. First, we trained a
matched-capacity attentive-probe model with the same seven video routes and
binding architecture, but with all video features replaced by zeros during both
training and prediction. This dummy-video control obtains 0.7448 on weighted-F1,
well below the corresponding attentive-probe baseline. Second, we kept the final
five-fold EEGBind ensemble fixed and zeroed the video routes only at candidate
inference time. This counterfactual obtains 0.7460 on weighted-F1. These controls
indicate that the gain of EEGBind is not explained by adding parameters alone
and that the trained model uses video-context routes in a measurable way, while
the source decision remains organized around EEG. The last two rows of
Table~\ref{tab:official_results} summarize these post-evaluation controls.

\subsection{Implementation Details}

Table~\ref{tab:implementation} summarizes the configuration used by the
five-fold EEGBind ensemble. EEGBind uses
ST-EEGFormer-large~\cite{yang2026steegformer} as the EEG backbone, seven
synchronized video-context routes from the official feature
release~\cite{neuromm2026baseline}, and an attentive probe for EEG-centric
binding. The EEG backbone is first initialized from our Task 1 fine-tuned
ST-EEGFormer checkpoint and then fine-tuned on Track 3 source labels before
multimodal binding. The repair stage is initialized from the binding model and
runs for one epoch with cross-entropy and binding losses disabled.

\begin{table}[t]
  \centering
  \caption{Main implementation settings for EEGBind.}
  \label{tab:implementation}
  \small
  \begin{tabular}{@{}p{0.36\linewidth}p{0.56\linewidth}@{}}
    \toprule
    Component & Setting \\
    \midrule
    EEG backbone & ST-EEGFormer-large initialized from a Task 1 fine-tuned checkpoint \\
    Track 3 EEG fine-tuning & 10 epochs, LR $5\times10^{-4}$, batch size 16 \\
    Optimizer / precision & AdamW, weight decay $10^{-4}$, class weights, fp16 AMP \\
    Video feature source & released seven-video feature cache \\
    Binding module & 2-layer, 8-head attentive probe with token/route dropout \\
    Binding training & 8 epochs, LR $3\times10^{-4}$, batch size 256 \\
    Repair stage & 1 epoch, LR $2\times10^{-5}$, batch size 256 \\
    Repair perturbations & Bernoulli route dropout, $p=0.15$, at least 3 routes retained; 1--2 masked routes, at least 5 retained \\
    Repair objectives & view-drop consistency, feature-target consistency, masked-route prediction, weak queue contrast \\
    Repair loss weights & 0.25, 0.02, 0.005, and 0.0005, respectively \\
    Test-time ensemble & subject-disjoint five-fold log-probability average \\
    \bottomrule
  \end{tabular}
\end{table}

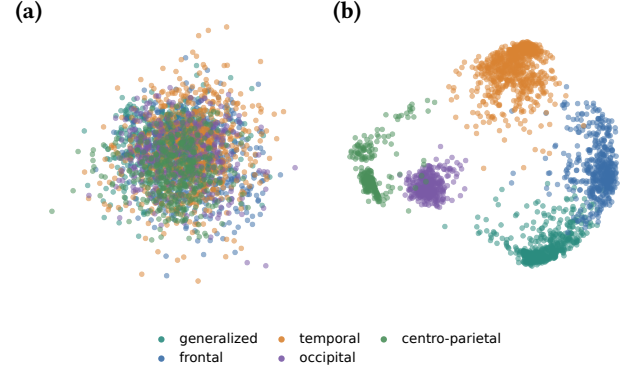
\begin{figure}[t]
  \centering
  \begin{tikzpicture}
    \node[anchor=south west, inner sep=0] (img) at (0,0)
      {\includegraphics[width=\linewidth]{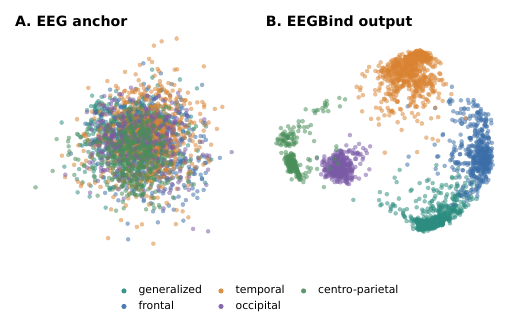}};
    \begin{scope}[x={(img.south east)},y={(img.north west)}]
      \fill[white] (0.006,0.908) rectangle (0.365,0.962);
      \fill[white] (0.501,0.908) rectangle (0.965,0.962);
      \node[anchor=north west, font=\bfseries, fill=white, inner sep=0.4pt] at (0.010,0.950) {(a)};
      \node[anchor=north west, font=\bfseries, fill=white, inner sep=0.4pt] at (0.505,0.950) {(b)};
    \end{scope}
  \end{tikzpicture}
  \par\vspace{-10pt}
  \caption{Out-of-fold representation map comparing the ST-EEGFormer feature
  space with the EEGBind output space on subject-disjoint validation folds.}
  \Description{A two-panel scatter plot comparing PCA projections of EEG anchor
  features and EEGBind output representations. The EEG anchor panel shows
  overlapped source classes, while the EEGBind output panel shows more separated
  class regions.}
  \label{fig:embedding_map}
\end{figure}

We further inspect the learned class geometry on the subject-disjoint
out-of-fold validation predictions. Figure~\ref{fig:embedding_map} compares a
two-dimensional principal component analysis (PCA) projection of the
ST-EEGFormer EEG anchor features with the corresponding EEGBind output space.
The EEG anchor features remain substantially overlapped across source classes,
whereas the EEGBind output space forms more separated class regions. This
projection is computed from subject-disjoint out-of-fold validation
predictions; Table~\ref{tab:official_results} reports the official hidden-set
weighted-F1.

\begin{figure}[t]
  \centering
  \includegraphics[width=\linewidth]{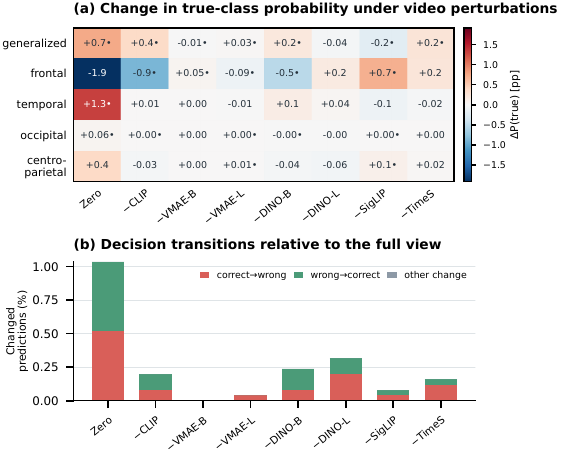}
   \par\vspace{-10pt}
  \caption{Out-of-fold video-perturbation analysis over 2,514 examples from 52
  subjects. (a) Mean classwise change in true-class probability after zeroing
  all video features or removing one route; dots denote subject-clustered 95\%
  bootstrap confidence intervals excluding zero. (b) Prediction transitions
  relative to the full view.}
  \Description{A two-panel out-of-fold perturbation figure. The upper panel is
  a five-by-eight heatmap showing changes in true-class probability for five
  source-level IED classes under zero-video and seven leave-one-route-out
  conditions. The lower panel is a stacked bar chart showing correct-to-wrong
  and wrong-to-correct prediction transitions. Zero-video changes about one
  percent of decisions, whereas every single-route removal changes fewer than
  one third of one percent.}
  \label{fig:video_type_patterns}
\end{figure}

Because EEGBind uses video as auxiliary context rather than as a direct source
classifier, we quantify its sensitivity to missing video information on all
2,514 positive examples from the subject-disjoint out-of-fold validation sets.
Figure~\ref{fig:video_type_patterns} compares the full view with a zero-video
view and seven leave-one-route-out views. Zeroing all routes changes 1.03\% of
the predictions, with equal proportions changing from correct to incorrect and
from incorrect to correct (0.52\% each). In contrast, removing any single route
retains 99.68--100\% prediction agreement with the full view. The classwise
confidence shifts further show that video context can affect individual
decisions, while the high single-route agreement indicates that no individual
video encoder dominates the source-level prediction.

\begin{figure}[!t]
  \centering
  \includegraphics[width=\linewidth]{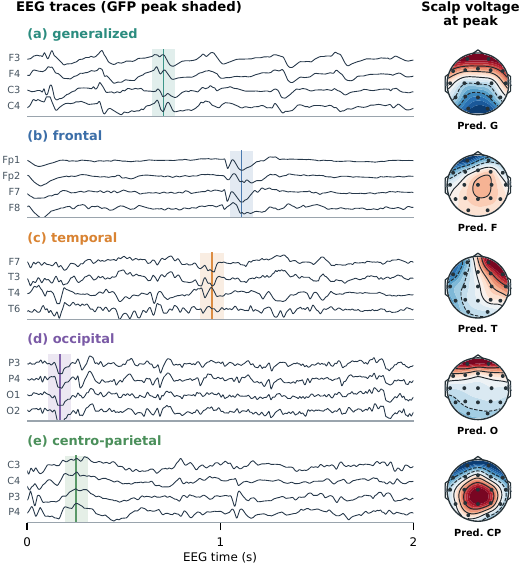}
  \caption{Representative out-of-fold EEG spatial cases for the five source
  types. Each row shows region-relevant EEG traces and the scalp voltage
  topography at the marked global-field-power peak. Cases are selected by
  proximity to their class EEG-embedding centroid among correctly classified
  examples. Topographies use a common-average reference and are normalized per
  case; blue and red denote negative and positive voltage, respectively.}
  \label{fig:single_case}
  \Description{A five-row qualitative EEG figure for generalized, frontal,
  temporal, occipital, and centro-parietal IEDs. Each row contains four EEG
  traces with a shaded global-field-power peak and a scalp voltage map computed
  at the same instant. The generalized map is broadly distributed, the temporal
  map shows a lateral field gradient, and the centro-parietal map has a central
  field maximum.}
\end{figure}

Figure~\ref{fig:single_case} complements the population-level perturbation
analysis with representative EEG spatial cases. For each source class, we select
the correctly classified out-of-fold example closest to its class centroid in
the normalized EEG embedding space. The shaded interval and vertical line mark
the maximum smoothed global-field-power peak, and the adjacent scalp map shows
the common-average-referenced voltage at that same instant. The maps are
normalized within each case to emphasize spatial field structure. They
visualize sensor-space field structure rather than inverse source estimates.

\subsection{Repair Analysis}

The repair variants in Table~\ref{tab:official_results} provide
official-evaluation evidence for the CE-free repair design. Adding
feature-target consistency improves the view-drop attentive-probe baseline from
0.8131 to 0.8203. Adding masked-route latent prediction further reaches 0.8239,
and adding the weak queue-based contrastive term reaches 0.8282. These results
support the repair stage as a lightweight robustness step around the learned
binding representation.

The five-fold ensemble applies the same EEG-centric binding and repair pipeline
to each subject-disjoint fold and averages their predictions. It improves over
the repaired single model without changing the source-classification objective.

\section{Discussion}

EEGBind assigns distinct roles to EEG and video. In Track 3, the source labels
are defined by EEG source-region
categories~\cite{neuromm2026challenge}, whereas synchronized video mainly
provides contextual evidence about the recording condition and patient state.
EEGBind therefore does not treat video features as direct evidence for
epileptogenic source. Instead, video-context routes are bound around the
EEG-centric representation to support source-level classification, while the
short repair stage improves consistency across route perturbations without
introducing another supervised update to the learned source-class boundary.

The post-evaluation controls examine two properties of the video pathway.
Dummy-video training isolates the contribution of attentive-probe capacity,
whereas zero-video inference measures the dependence of a trained model on its
video routes under the challenge protocol.

Track 3 differs from the binary NeuroMM tasks by shifting the objective from
detecting or ranking IED candidates to discriminating among five source regions.
Its errors therefore include confusions between anatomically different source
categories in addition to missed or spurious IED decisions. The EEGBind pipeline
accordingly keeps EEG as the primary modality and uses multimodal context as
auxiliary evidence rather than as an unconstrained fusion signal.

Overall, the official evaluation results support the EEGBind pipeline:
EEG-centric multimodal binding, lightweight CE-free repair,
and subject-disjoint ensembling. Future work should build stronger
subject-heldout validation for source labels and study bounded ways to use
multimodal context without disrupting source-sensitive EEG decisions.

\section{Conclusion}

We presented EEGBind, an EEG-centric multimodal binding framework for Track 3
NMM-Source-IED, where the key challenge is to use synchronized video context
without perturbing the EEG representation that carries the source-region
evidence. Its binding mechanism treats EEG as the source anchor and
binds video as bounded candidate context, while CE-free view-consistent repair
improves robustness under route perturbations. In the official evaluation,
Multimodal binding improves the
EEG-only baseline from 0.6444 to 0.8019, repair further raises the single-model
score to 0.8282, and the subject-disjoint five-fold ensemble reaches 0.8395 on
weighted-F1. Post-evaluation controls with dummy-video training and zero-video
inference drop to 0.7448 and 0.7460, indicating that the gain is not explained
by model capacity alone but by meaningful video-context binding. These results
support this design as a practical strategy for source-level IED classification.

\begin{acks}
This work is partially supported by the Guangdong Basic and Applied Basic Research Foundation
(2026A1515011793), the Youth S\&T Talent Support Programme of Guangdong Provincial Association for Science and Technology (SKXRC2025467), and the Research Travel Grant of the Base of
Red Bird MPhil (RBM) at The Hong Kong University of Science and Technology (Guangzhou). 
We would like to thank our RBM Project Supervisor Ning LI for the academic support.

Figure~\ref{fig:method_overview} includes
AI-generated illustrative elements created with GPT Image 2. They were used only
for visual illustration; all scientific claims, methods, results, plots, and
reported values were produced by the authors.
\end{acks}

\clearpage
\balance
\bibliographystyle{ACM-Reference-Format}
\bibliography{references}

\end{document}